\documentclass[10pt,twocolumn,letterpaper]{article}

\usepackage{cvpr}              
\definecolor{cvprblue}{rgb}{0.21,0.49,0.74}
\usepackage[pagebackref,breaklinks,colorlinks,allcolors=cvprblue]{hyperref}
\usepackage{multirow}
\usepackage{multicol}
\usepackage{makecell}
\usepackage{amsmath,amssymb}
\usepackage{booktabs}
\usepackage{array}
\usepackage{hhline}
\usepackage{float}
\usepackage[table]{xcolor}

\usepackage{stfloats}

\title{AnyExperts: On-Demand Expert Allocation for Multimodal Language Models with Mixture of Experts}

\author{
  Yuting Gao$^{1}$\thanks{The first three authors contributed equally. This work was done when H. Zhao was an intern at AntGroup.}
  \quad Lan Wang$^{1,*}$
  \quad Hengyuan Zhao$^{1,2,*}$
  \quad Linjiang Huang$^{2}$
  \quad Si Liu$^{2}$
  \quad Qingpei Guo$^{1}$ \\[0.25cm]
  \small
  $^1$AntGroup \quad
  $^2$Beijing University of Aeronautics and Astronautics \\[0.25cm]
  \tt\small yutinggao.sh@gmail.com
}

\begin{document}
\maketitle
\begin{abstract}
Multimodal Mixture-of-Experts (MoE) models offer a promising path toward scalable and efficient large vision-language systems. However, existing approaches rely on rigid routing strategies (typically activating a fixed number of experts per token) ignoring the inherent heterogeneity in semantic importance across modalities. This leads to suboptimal compute allocation, where redundant tokens consume as many resources as critical ones. To address this, we propose \textbf{AnyExperts}, a novel \textbf{on-demand, budget-aware dynamic routing} framework that allocates a variable total number of expert slots per token based on its semantic importance. Crucially, to prevent uncontrolled compute growth, the total slots per token are constrained within a fixed range, and each slot is filled by either a real expert or a virtual expert, with the virtual share capped at a small maximum (\textit{e.g.}, 20\%). The model then adaptively balances the real-to-virtual ratio per token, assigning more real experts to semantically rich regions and relying more on virtual experts for redundant content. Evaluated across diverse tasks in visual understanding, audio understanding, and NLP understanding, AnyExperts improves performance under the same compute budget. Notably, on general image/video tasks, it achieves comparable accuracy with 40\% fewer real expert activations; on text-dense tasks (OCR and NLP), it maintains performance while reducing real expert usage by 10\%. These results demonstrate that fine-grained, importance-driven expert allocation significantly enhances both the efficiency and effectiveness of multimodal MoE models.
\end{abstract}    
\section{Introduction}
\label{sec:intro}

Multimodal large language models (MLLMs) ~\cite{yin2024survey,zhang2024mm,lin2024video,li2024llava,wang2024qwen2,bai2025qwen2} have demonstrated remarkable capabilities in complex reasoning by effectively integrating multiple modalities. However, their ever-increasing parameter counts have led to substantial computational overhead. To address this challenge, the Mixture-of-Experts (MoE) architecture has emerged as a promising paradigm for efficient scaling: by sparsely activating a subset of expert subnetworks per input, MoE enables significant parameter growth while maintaining controlled computational cost~\cite{team2025kimi, ai2025mingflashomni, 2025qwen3-omni} compared to traditional dense models~\cite{dai2024deepseekmoe,li2022sparse}.

\begin{figure}[t]
    \centering
    \includegraphics[width=\linewidth]{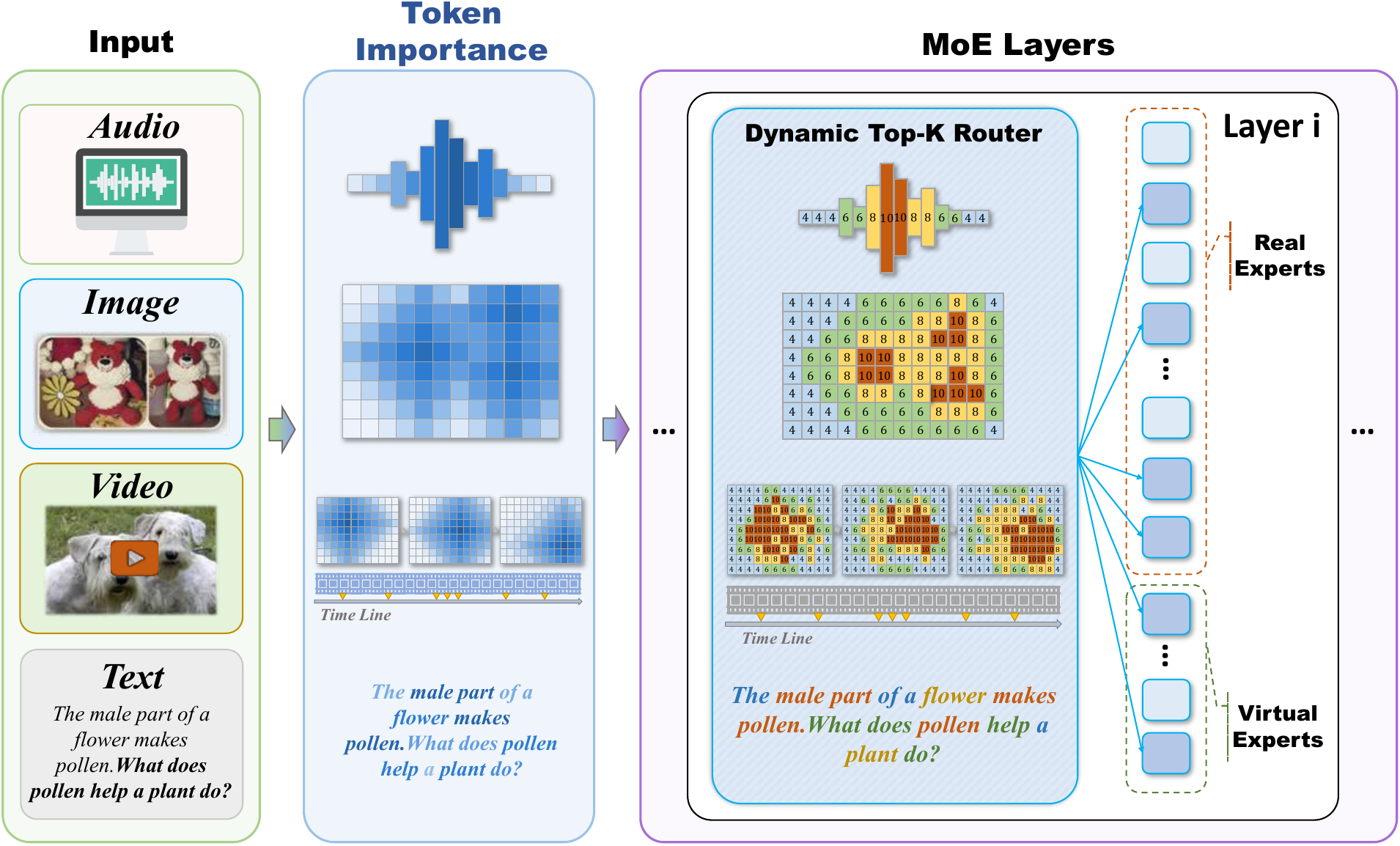}
    \caption{AnyExperts replaces rigid Top-$K$ routing with importance driven expert allocation: tokens from multimodal inputs are assigned a dynamic number of expert slots (composed of real and virtual experts), based on their semantic significance. Informative tokens activate more real experts; redundant ones rely more on virtual computation, ensuring compute is focused precisely where it matters most.}
    \label{fig:anyexperts_demo}
\end{figure}

However, existing MoE approaches in multimodal settings remain limited by rigid routing strategies. Most notably, they enforce a fixed number $K$ of activated experts per token, irrespective of the input’s modality or semantic content. This \textbf{one-size-fits-all} allocation fails to account for the inherent heterogeneity across modalities: visual or video inputs often contain substantial redundancy (\textit{e.g.}, homogeneous background regions), whereas textual tokens (such as name entities or descriptive phrases) tend to carry denser information. Consequently, static Top-$K$ routing indiscriminately assigns the same computational budget to both highly informative and redundant tokens, leading to suboptimal efficiency and representation quality.

While recent efforts have explored dynamic expert activation in unimodal language models, such as allocating more experts to harder tasks via probability thresholds (\textit{e.g.}, Top-$P$~\cite{huang2024harder}) or skipping computation for tokens deemed uninformative (\textit{e.g.}, MoE++~\cite{jin2024moe++}), these strategies do not readily generalize to multimodal settings. One limitation lies in the Top-$P$ method: when processing data with background or high levels of redundant visual information, this method causes redundant visual tokens to also activate a large number of experts, which in turn leads the model to waste computational resources on uninformative regions. Another issue arises with strategies like MoE++, which simplify resource allocation into a decision of "normal computation, reduced computation, or skipped computation". However, they still operate on a fixed number of experts. For high-value tokens (\textit{e.g.}, partially occluded objects), which require more (rather than fewer) expert processing to achieve effective cross-modal reasoning, the aforementioned strategies fail to meet their differentiated capacity needs. These issues underscore the need for a multimodal-aware routing mechanism that estimates semantic importance independently of gating confidence and supports both up-scaling and down-scaling of expert allocation per token.

In this work, we present \textit{AnyExperts}, a dynamic expert allocation framework that enables \textbf{on-demand} routing with strict computational control. The core idea is to assign each multimodal token a \textbf{dynamically determined total number of expert slots}, proportional to its estimated semantic importance. This importance score is predicted by a lightweight MLP that outputs a continuous value per token and is trained end-to-end without external supervision. To avoid uncontrolled compute growth and training instability caused by over-activation, this total is constrained within a range, ensuring the overall computation remains comparable to a standard top-$K$ baseline. Within this budget, each token allocates its slots between real experts and virtual experts, with the virtual expert’s share capped at a small maximum (\textit{e.g.}, 20\%). The model then adaptively adjusts the real-to-virtual ratio per token: highly informative tokens (\textit{e.g.}, object regions or named entities) prioritize real experts, while redundant tokens rely more on the virtual expert. This design shifts MoE routing from rigid, token-agnostic allocation to \textbf{budget-aware, token-specific resource assignment that concentrates real computation precisely where it matters most}.

We evaluate \textit{AnyExperts} across a diverse set of multimodal tasks, encompassing visual understanding (general image-text question answering, document understanding, and video question answering) and audio understanding (\textit{e.g.}, automatic speech recognition), and language understanding. At inference time, it enables flexible trade-offs between computation and performance: when using the same average number of activated experts per token as a standard top-$K$ baseline, \textit{AnyExperts} achieves higher scores thanks to its importance-aware routing; conversely, it can reduce the average expert count while maintaining performance parity. These results demonstrate that the efficiency of Mixture-of-Experts models can be substantially enhanced by dynamically allocating computation in accordance with the semantic importance of individual tokens.

Our core contributions can be summarized as:
\begin{itemize}
    \item We propose AnyExperts, a novel on-demand, dynamic routing framework for multimodal Mixture-of-Experts models, which allocates a variable number of experts per token based on its semantic importance. 
    \item AnyExperts consistently boosts performance under the same expert budget, and reduces compute at parity, across diverse tasks, demonstrating that importance-driven allocation maximizes the utility of every expert.
\end{itemize}
\section{Related Works}
\label{sec:related_works}

\begin{figure*}[t]
    \centering
    \includegraphics[width=\linewidth]{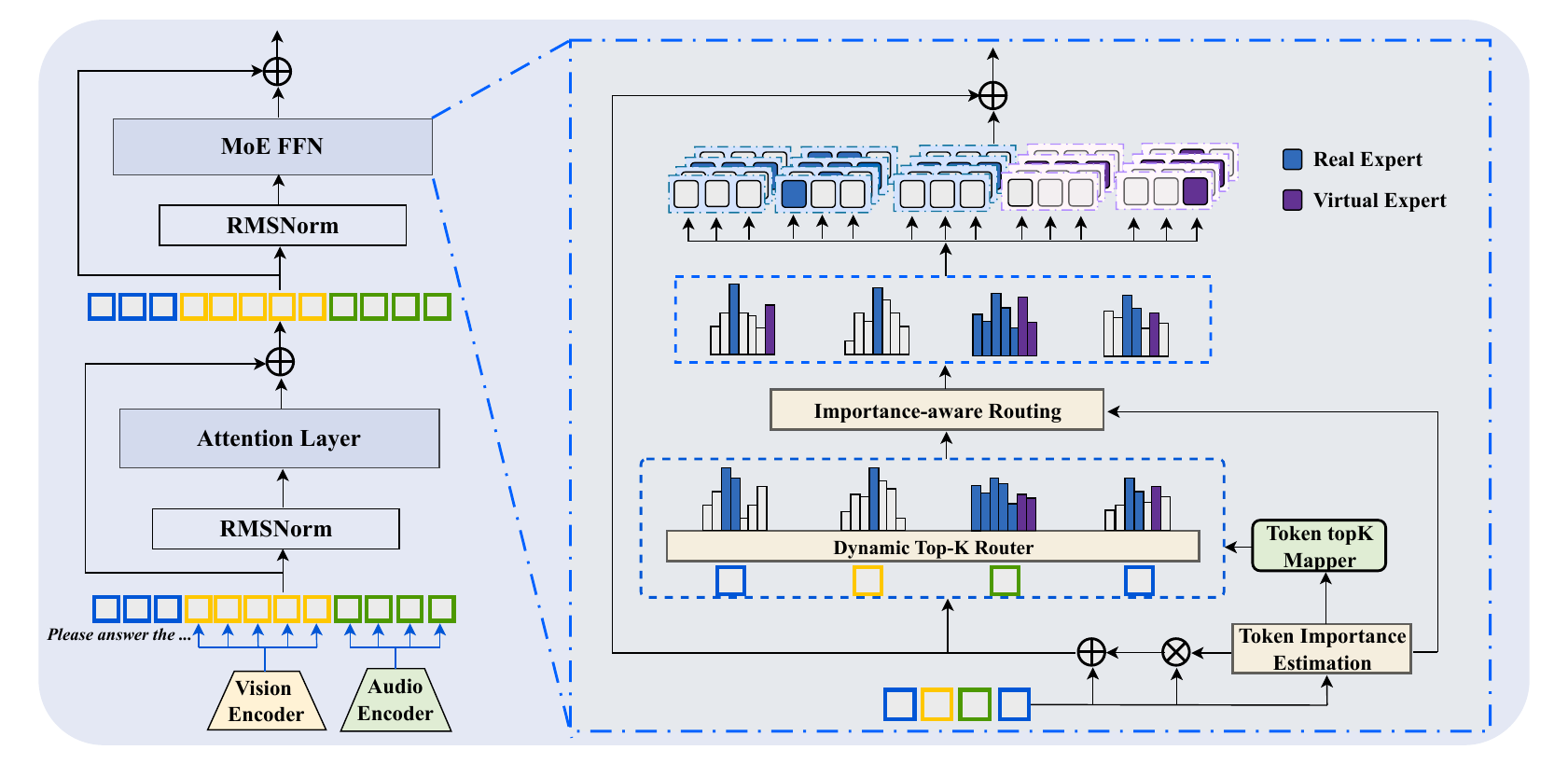}
    \caption{Overview framework of AnyExperts. For each input token, a lightweight gating network estimates an importance score that dynamically determines the total number of expert. The router then allocates these slots between real experts and virtual expert, enabling \textit{per-token adaptive expert activation}—\textit{i.e.}, semantically rich tokens activate more real experts, while redundant ones use fewer, achieving fine-grained, budget-aware computation.}
    \label{fig:main}
\end{figure*}

\subsection{Static Routing MoE Architectures}
Since models such as GShard~\cite{lepikhin2020gshard} pioneeringly applied MoE to translation tasks for large-scale language models, the MoE architecture has demonstrated strong capabilities. The milestone Switch Transformers~\cite{fedus2022switch}, by extending the MoE method, achieved unprecedented model scale and training efficiency in visual understanding tasks.
Most routing mechanisms in MoE models assign a fixed number of experts to each token. Among these, top-$K$ routing~\cite{shazeer2017outrageously,jiang2024mixtral,yang2025qwen3} is the most widely adopted approach. It computes a gating score reflecting each expert’s suitability for the input token, then selects the top-$K$ experts with the highest scores for activation. 
Subsequent work~\cite{zoph2022st,lepikhin2020gshard,fedus2022switch,wu2024multi} has primarily focused on balancing token assignments across experts under the constraint of fixed top-$K$ routing. Other efforts have extended MoE models to multimodal domains: for instance, Uni-MoE~\cite{li2025uni} supports speech, image-text, and video inputs, while DeepSeek-MoE~\cite{dai2024deepseekmoe} and XMoE~\cite{yang2024xmoe} introduce fine-grained expert segmentation into the MoE architecture. Despite these advances, all aforementioned approaches adhere to the same core design principle: activating a fixed number of experts per token.

\subsection{Dynamic Routing Approach}

Recognizing that tokens vary in semantic importance and computational demand, a line of work has explored dynamic expert activation, where the number of experts per token is adapted based on input characteristics.  Top-$P$ routing~\cite{huang2024harder} activates a variable set of experts by accumulating sorted gating scores until a predefined probability threshold is reached, effectively allocating more experts to harder examples. DA-MoE~\cite{aghdam2024moe} estimates token importance via attention weights in transformer layers and dynamically selects the number of experts for language tokens. Other approaches achieve dynamic behavior through heterogeneous expert designs: MoE++~\cite{jin2024moe++} introduces specialized expert types to reduce resource usage for less informative tokens. Most extremely, LongCat-Flash~\cite{LongCat} employs only zero-computation experts, allowing the effective number of activated experts to vary per token while maintaining low compuational cost.

However, these approaches are not well aligned with multimodal scenario: threshold-based methods like Top-$P$ generate the same routing confidence for regions with high and low semantic importance, and thus allocate excessive experts to regions with low semantic importance, wasting computational resources, while skip-oriented designs such as MoE++ only reduce computation for uninformative tokens and lack the capacity to allocate more experts to critical regions that demand enhanced modeling. This highlights the need for a multimodal routing mechanism that assesses semantic importance independently of gating confidence and allows expert capacity to be scaled up or down per token as needed.

\section{Methods}
\label{sec:methods}

In this section, building upon the motivation outlined above, we present the technical details of AnyExperts, a dynamic expert allocation framework designed to overcome the inefficiencies of rigid, one-size-fits-all token routing. As illustrated in Figure~\ref{fig:main}, AnyExperts first estimates an importance score for each token and maps this score to a target number of activated experts, allocating more experts to semantically rich tokens and fewer to redundant ones. To enable flexible and efficient routing, we introduce a \textit{virtual expert}, a lightweight routing target that bypasses standard expert computation and directly maps the input token representation, thereby incurring negligible computational cost.

\subsection{Token Importance Estimation}
\label{sec:token_importance}

To address the mismatch between routing confidence and true semantic importance in multimodal inputs, we introduce a dedicated importance estimation mechanism that operates independently of the MoE gating network. Specifically, we treat tokens from all modalities as equal and employ a single lightweight MLP to autonomously judge the importance of tokens across different modalities. For any token \(i\) with hidden state \(\mathbf{h}_i \in \mathbb{R}^d\), the MLP module outputs an importance score informed by the token's contextual distribution: 

\begin{equation}\label{eq:importance_score}
    s_i = \text{MLP}(\text{LayerNorm}(x_i)),
\end{equation}
where \(s_i\) serves as a key determinant for subsequent expert allocation. To ensure non-negativity and appropriate scaling, we map \(s_i\) to an importance weight \(w_i\) via a sigmoid activation:
\begin{equation}
    w_i = \text{Sigmoid}(s_i),
\end{equation}
we then scale the weight by a positive control parameter \(\alpha \in \mathbb{R}^+\) (typically $\alpha=0.01$) and apply it to the original hidden state \(\mathbf{h}_i\) via a residual connection:
\begin{equation}\label{eq:residual_fusion}
    \mathbf{h}_i' = \mathbf{h}_i + \alpha \cdot w_i \cdot \mathbf{h}_i
\end{equation}
where \(\mathbf{h}_i'\) denotes the updated hidden state after fusion. This design amplifies features of high-importance tokens during forward propagation and provides a distinct supervision signal for the MLP, enabling it to refine its importance estimation (including distinguishing importance across different modalities) through end-to-end backpropagation.

\subsection{Dynamic Allocation with Virtual Reserve}
\label{sec:dynamic_allocation}

In this section, we present a dynamic expert allocation framework enabling on-demand routing with computational control. 

\paragraph{Dynamic Expert Count}

We propose a dynamic expert allocation mechanism that assigns each token a variable number of expert slots, proportional to its semantic importance as quantified by the importance score introduced in Section~\ref{sec:token_importance}. To avoid uncontrolled compute growth and training instability from over-activation, the total slots are constrained within $[K_{\min}, K_{\max}]$, keeping overall computation comparable to a standard top-$K$ baseline.  

Using token $i$'s importance score $w_i$, the total experts $\hat{K}_i$ for token $i$ is dynamically determined within the interval $[K_{\min}, K_{\max}]$ as follows:  
\begin{equation}
    \hat{K}_i = K_{\min} + (K_{\max} - K_{\min}) \cdot w_i,
\end{equation}

This total comprises both real experts ($K_i^{\text{real}}$) and virtual experts ($K_i^{\text{virtual}}$), such that $\hat{K}_i = K_i^{\text{real}} + K_i^{\text{virtual}}$. To mitigate overreliance on virtual experts, their contribution is constrained to at most a fraction $\rho_{\max}$ (typically $\rho_{\max}=0.2$).

\paragraph{Importance-aware Routing} To concentrate real computational resources on semantically critical tokens, we introduce importance-aware routing weights that modulate each token’s affinity towards real versus virtual experts. Let $r_{i,e} \in \mathbb{R}^{\hat{K}_i}$ denote the raw routing logit output by the routing network for token $i$ and expert $e$, where $e$ ranges over the full set of experts $\mathcal{E} = \mathcal{E}^{\text{real}} \cup \mathcal{E}^{\text{virtual}}$.

For token $i$ with importance weight $w_i$, we define importance-modulated routing weights as:

\begin{align}\label{eq:import_routing}
    \phi_i^{\text{real}} &= 1 + \alpha \cdot w_i, \quad e \in \mathcal{E}^{\text{real}}, \\
    \phi_i^{\text{virtual}} &= 1 - \alpha \cdot w_i, \quad e \in \mathcal{E}^{\text{virtual}},
\end{align}
where $\alpha$ controls the strength of the importance-induced bias. This encourages high-importance tokens (\textit{e.g.}, salient object regions) to prefer real experts, while redundant tokens are steered toward virtual ones. 

The refined routing scores $r_{i,e}'$ are obtained by reweighting the raw logits:

\begin{equation}\label{eq:importance_aware_modulation}
    r_{i,e}' = 
    \begin{cases} 
    r_{i,e} \cdot \phi_i^{\text{real}} & e \in \mathcal{E}^{\text{real}}, \\
    r_{i,e} \cdot \phi_i^{\text{virtual}} & e \in \mathcal{E}^{\text{virtual}}.
    \end{cases}
\end{equation}

The top-$\hat{K}_i$ experts are then selected on $r_{i,e}'$, forming the active expert set $\mathcal{E}_i$. For each selected expert $e \in \mathcal{E}_i$, if $e$ is a real expert, its output is computed via standard forward pass; if $e$ is a virtual expert, it bypasses computation and directly returns the input hidden state as output (\textit{i.e.}, identity mapping). The final normalized combination weights are computed as:
\begin{equation}
    \gamma_{i,e} = \frac{\sigma(r_{i,e}')}{\sum_{e \in \mathcal{E}_i} \sigma(r_{i,e}') + \epsilon} \cdot \lambda,
\end{equation}
where $\epsilon = 10^{-8}$ ensures numerical stability and $\lambda$ scales the overall contribution magnitude.

This importance-aware routing strategy transforms expert assignment from a rigid, token-agnostic process into a budget-conscious, token-specific mechanism, enabling on-demand allocation of computational resources.
\subsection{Training Objective}
In addition to the standard next-token prediction loss for language modeling, we introduce two auxiliary losses to enhance the stability and effectiveness of our dynamic expert allocation framework. The first is a token importance regularization loss, which stabilizes the estimation of token importance scores produced by the \(\mathtt{MLP}\) (section~\ref{sec:token_importance}), ensuring they remain reliable and semantically meaningful. The second is a balancing control loss, which encourages equitable utilization of experts and prevents degenerate routing behaviors. Together, these auxiliary losses reinforce the consistency, robustness, and efficiency of the importance-aware routing mechanism.

\paragraph{Token Importance Score Regularization}
To stabilize the importance estimation of the \(\mathtt{MLP}\) and prevent over-confident or noisy outputs, we introduce a token importance score regularization (TIR) loss on the raw importance score \(s_i\). Recall that \(s_i\) is passed through a sigmoid to yield importance weights \(w_i = \sigma(s_i)\), which are scaled by $\alpha$ for dynamic expert allocation. To constrain the magnitude of $s_i$, we define: 
\begin{equation}
    \mathcal{L}_{\text{TIR}} = \frac{1}{N}\sum_{n=1}^{N} (w_{i})^{2}
\end{equation}  

where \(N\) denotes the total token in the batch. This regularization loss prevents most \(s_i\) from growing excessively large, thereby limiting the number of activated experts to a reasonable range, reducing computational cost, and promoting stable, balanced importance estimates. 

\paragraph{Balancing Control}

Balancing expert utilization in AnyExperts is non-trivial due to the coexistence of real and virtual experts. Virtual experts (implemented as $E_{\text{virtual}}$ identical routing copies) can be selected frequently, yet treating them as independent experts would either dilute the balancing signal (if included naively) or induce instability (if excluded entirely). 

To resolve this, we introduce a unified load accounting scheme: all tokens routed to virtual experts are aggregated into a total count $T_{\text{virtual}}$, which is then evenly distributed across the $E_{\text{virtual}}$ copies as $T_{\text{virtual}}/E_{\text{virtual}}$, while real experts retain their exact token assignment counts. 

Let $E_{\text{real}}$ denote the number of real experts, and define the total number of expert as  $E = E_{\text{real}} + E_{\text{virtual}}.$ This calibrated load vector underpins global balancing loss that enforces equitable utilization across all $E$ expert slots:
\begin{equation}
    \mathcal{L}_{\text{Balance}} = \sum_{k=1}^{E} f_k \cdot p_k,
\end{equation}

where $p_k = \frac{1}{N} \sum_{i=1}^N P(e = k \mid x_i)$ is the mean routing probability assigned to expert $k$, and the calibrated token fraction $f_k$ is defined as:
\[
f_k = 
\begin{cases}
\dfrac{c_k}{N}, & 1 \leq k \leq E_{\text{real}}, \\
\dfrac{T_{\text{virtual}}}{E_{\text{virtual}} \cdot N}, & E_{\text{real}} < k \leq E,
\end{cases}
\]

with $c_k$ denoting the raw token count for expert $k$, and $N$ the total number of tokens in the batch. By smoothing virtual expert loads while preserving real expert fidelity, this formulation ensures efficient, noise-robust dynamic allocation without compromising routing expressivity.

\paragraph{Total Loss}
The total training objective integrates three components:
\begin{equation}
    \mathcal{L} = \mathcal{L}_{\text{LM}} + \lambda_{tir} \mathcal{L}_{\text{TIR}} + \lambda_{bal} \mathcal{L}_{\text{Balance}},
\end{equation} 
where \(\mathcal{L}_{\text{LM}}\) is the standard next-token cross-entropy loss that drives core language modeling performance, $\mathcal{L}_{\text{TIR}}$ regularizes the raw importance logits to yield stable and well-calibrated token importance scores, and $\mathcal{L}_{\text{Balance}}$ enforces balanced usage between real and virtual experts to avoid routing degeneracy. The small coefficients $\lambda_{tir}$ and $\lambda_{bal}$ are set to 0.001 and 0.01 respectively, ensure that the auxiliary losses support the primary language modeling objective (without overpowering). This joint optimization enables robust dynamic expert allocation while preserving strong generative capabilities.

\section{Experiments}
\label{sec:experiments}

\begin{table*}[t]
\centering
\small
\setlength{\tabcolsep}{4.0pt} 
\caption{Performance of our proposed AnyExperts  across multimodal benchmarks.}
\label{tab:our_results}
\begin{tabular}{l c | l c | l c | l c | l c}
\toprule
\multicolumn{2}{c|}{\textbf{Image Understanding $ \uparrow $}} & 
\multicolumn{2}{c|}{\textbf{OCR $ \uparrow $}} & 
\multicolumn{2}{c|}{\textbf{NLP $ \uparrow $}} & 
\multicolumn{2}{c|}{\textbf{Speech $ \downarrow $}} & 
\multicolumn{2}{c}{\textbf{Video $ \uparrow $}} \\
\midrule

MMBench           & 79.73 & ChartQA            & 83.04 & ARC-C   & 90.17 & Aishell1  & 1.70  & MVBench               & 66.40 \\
AI2D             & 81.19 & TextVQA             & 73.42 & GSM8k   & 84.76& Aishell2-test-ios & 2.78 & VideoMME              & 60.59 \\
MMMU   & 49.11 & OCRBench           & 85.10  & GPQA   & 41.92 &  Aishell2-test-android & 2.76 & OvOBench        & 44.83 \\
MMVet         & 66.15 &                    &      &           &      & Librispeech-test-other & 3.01 &                       &      \\
MathVista  & 66.70 &                    &      &           &      & Librispeech-test-clean & 1.38 &                       &      \\
MMStar           & 60.48 &             &  &    &  &   &  &                &  \\
\midrule
\rowcolor{gray!20} Average & 67.22 & Average & 80.52 & Average & 72.28 & Average & 2.32 & Average & 57.27 \\
\bottomrule
\end{tabular}
\end{table*}

\subsection{Experimental Setup}

\subsubsection{Training Details}

Our final results are obtained by following the understanding training protocol of Ming-Omni~\cite{ming-omni}. This entire pipeline comprises two pretraining stages and two subsequent supervised fine-tuning stage. Consistent with Ming-Omni, the full training pipeline uses 1.1 trillion multimodal tokens during pretraining, followed by supervised fine-tuning on: 30 million image–text, 10 million video–text, 230 thousand hours of speech–text instruction-following samples. For ablation studies, we use a reduced-scale variant of the full pipeline, sampling approximately 1\% of the data from each stage (both in pretraining tokens and instruction-tuning examples) to enable efficient experimentation while preserving the overall data distribution.

\subsubsection{Model Configuration}
For our final model, we adopt Ling-mini-2.0~\cite{ling-mini} architecture as the base model. The original Ling-mini-2.0 employs 256 experts and activate exactly 8 experts per token in a uniform manner. In our AnyExperts framework, however, we dynamically adjust the number of activated experts per token to balance training stability, expert utilization efficiency, and computational cost. Specifically, during training, we set the minimum and maximum number of activated experts per token to $K_{min}=8, K_{max}=12$, and introduce $E_{virtual}=64$ virtual expert copies. This yields a virtual expert ratio of $\rho_{\max}=0.2$. Consequently, the expected number of real experts activated per token lies in the interval $[8 \times 0.8, 12 \times 0.8]=[6.4,9.6]$, with an average of approximately 8. As a result,  the total training time is on par with that of the baseline. 

At inference, we retain the same dynamic range $K_{min}=8$, $K_{max}=12$ and observe consistent performance gains. Notably, we reduce the average number of activated experts to 7.2 (90\% of the baseline’s fixed $K$=8) without any degradation in downstream performance, all metrics remain on par with the baseline. We attribute this robustness to the sufficient and stable training of individual experts under the AnyExperts regime, as each real expert has been adequately optimized through diverse routing patterns during training, enabling high-quality predictions even with fewer active experts at test time.

\subsubsection{Evaluation Benchmark}
\label{eval_benchmarks}
To comprehensively evaluate the capabilities of our proposed model, we conduct extensive experiments across five domains: general image–text question answering, optical character recognition (OCR), video understanding, audio understanding, and natural language processing (NLP). Our evaluation encompasses a diverse set of established benchmarks, including AI2D~\cite{hiippala2021ai2d}, MMBench~\cite{liu2024mmbench}, MMMU~\cite{yue2024mmmu}, MM-Vet~\cite{yu2023mm}, and MathVista~\cite{lu2023mathvista} for general QA; OCRBench~\cite{liu2024ocrbench}, ChartQA~\cite{masry2022chartqa}, and TextVQA~\cite{singh2019towards} for OCR; OVOBench~\cite{niu2025ovo}, MVBench~\cite{li2024mvbench}, and VideoMME~\cite{fu2025video} for video understanding; Aishell-1~\cite{bu2017aishell}, Aishell-2 (test-ios and test-android)~\cite{du2018aishell}, and LibriSpeech (test-clean and test-other)~\cite{panayotov2015librispeech} for audio understanding; and ARC-C~\cite{clark2018think}, GSM8K~\cite{cobbe2021training}, and GPQA~\cite{rein2024gpqa} for NLP.

\subsection{Main Results}

We present the main evaluation results of AnyExperts trained on full-scale multimodal data in Table~\ref{tab:our_results}. The model has 16B parameters total, with an average of 8.0 experts activated per token during inference, yielding an effective activated capacity of 1B parameters. Despite this moderate activated scale, AnyExperts achieves strong performance across diverse domains: 66.40\% on MVBench (video), 81.19\% on AI2D (diagram understanding), 83.04\% on ChartQA (OCR), and 90.17\% on ARC-C (NLP), demonstrating the effectiveness of our importance-aware routing when scaled to comprehensive training data. Notably, reducing average activation to 7.2 (a 10\% reduction) incurs no performance degradation, confirming that AnyExperts can flexibly trade off computation without sacrificing accuracy, detailed results are provided in the supplementary material.

\subsection{Ablation Studies}

In this section, we present ablation studies on several key components of our approach to analyze their individual contributions. Unless otherwise specified, the scores for General QA, OCR, Video, Audio, and NLP report the average performance across 5,3,3,5 and 3 benchmarks, as introduced in Section~\ref{eval_benchmarks}. Detailed results for each individual benchmark are provided in the Appendix.

\noindent\textbf{Importance-aware Hidden States} As described in Section~\ref{sec:token_importance}, after computing the importance weight for each token, we use it to modulate the corresponding hidden states with equation~\ref{eq:residual_fusion}. To assess the impact of this design, we ablate the importance-aware hidden state modulation and report results in Table~\ref{exp:reinforce}.  Its removal consistently degrades performance across all five evaluation domains, highlighting its critical contribution to model expressivity and the quality of the resulting representations.
 
\begin{table}[H]
\caption{Ablation study on the importance-aware hidden state modulation (HSM). }
\label{exp:reinforce}
\centering
\setlength{\tabcolsep}{4pt}
\resizebox{0.88\linewidth}{!}{
    \begin{tabular}{c|ccccc}
    \hline
    \multirow{2}{*}{\textbf{Setting}} &
    \multicolumn{5}{c}{\textbf{Accuracy on Tasks}} \\ \cline{2-6}
    & General$ \uparrow $ & OCR$ \uparrow $ & Video$ \uparrow $ & Audio$ \downarrow $ & NLP$ \uparrow $ \\    \hline
     \rowcolor{gray!20}\makecell{w HSM(Ours)} & 54.72 & 71.17 & 46.26 & 2.82 & 70.88 \\
     \makecell{w/o HSM} & 53.15 & 69.38 & 44.63 & 2.81 & 68.78 \\
    \hline
    \end{tabular}
}
\end{table}

\noindent\textbf{Impact of Importance-aware Routing} As detailed in Section~\ref{sec:dynamic_allocation} equation~\ref{eq:importance_aware_modulation}, we refine the router’s gating weights by incorporating expert importance scores to encourage more informed token-to-expert assignment. To assess the contribution of this routing mechanism, we ablate the importance-aware adjustment and report results in Table~\ref{exp:twf}. Removing this component leads to consistent performance drops, confirming that calibrating router weights with importance signals is crucial for effective sparse activation. 

\begin{table}[H]
\caption{Ablation study on the importance-aware routing (IAR).}
\label{exp:twf}
\centering
\setlength{\tabcolsep}{4pt}
\resizebox{0.88\linewidth}{!}{
    \begin{tabular}{c|ccccc}
    \hline
    \multirow{2}{*}{\textbf{Setting}} &
    \multicolumn{5}{c}{\textbf{Accuracy on Tasks}} \\ \cline{2-6}
    & General$ \uparrow $ & OCR$ \uparrow $ & Video$ \uparrow $ & Audio$ \downarrow $ & NLP$ \uparrow $ \\    \hline
    \rowcolor{gray!20}\makecell{w IAR(Ours)} & 54.72 & 71.17 & 46.26 & 2.82 & 70.88 \\
    \makecell{w/o IAR} & 52.46 & 69.95 & 42.36 & 2.83 & 69.88 \\
    \hline
    \end{tabular}
}
\end{table}

\begin{figure*}[htbp] 
\centering
\includegraphics[width=1.0\textwidth, keepaspectratio]{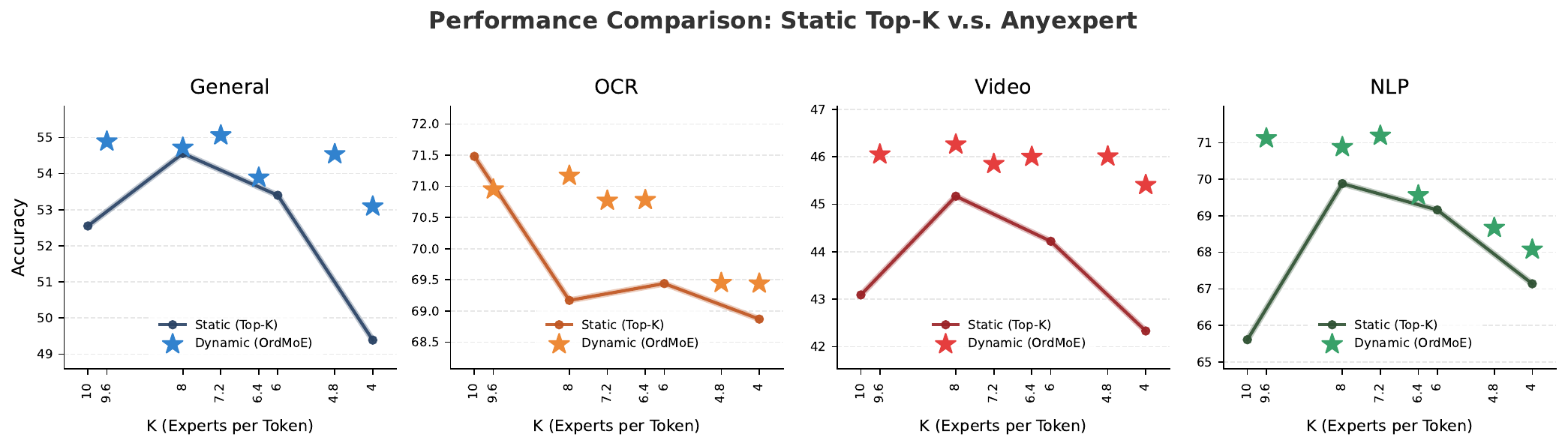}
\caption{Performance sensitivity to activated experts number under static v.s. dynamic routing. Results for general QA, video QA, OCR and NLP are shown; audio understanding exhibits minimal variation across budgets and is omitted for clarity, with full results provided in the supplementary material.}
\label{fig:expert_sensitivity}
\end{figure*}

\noindent\textbf{The choice of $\alpha$} As shown in Equation~\ref{eq:residual_fusion} in Section~\ref{sec:token_importance} and Equation~\ref{eq:import_routing} in Section~\ref{sec:dynamic_allocation}, the hyperparameter $\alpha$ controls the strength of importance-induced routing in AnyExperts. By default, we set $\alpha=0.01$. Table~\ref{exp:moralpha} presents an ablation over $\alpha \in \{0.005, 0.01, 0.05, 0.1\}$. Our default choice ($\alpha=0.01$, highlighted) consistently achieves the best performance across all five tasks. Slightly reducing $\alpha$ to 0.005  weakens the influence of token importance, leading to under-utilization of the dynamic routing mechanism. Conversely, increasing $\alpha$ to 0.05 or 0.1 amplifies the importance signal too aggressively, which perturbs the original embedding distribution and disrupts training stability, resulting in performance degradation (particularly noticeable in OCR and NLP tasks). This confirms that a moderate $\alpha$ (\textit{i.e.} 0.01) strikes the optimal balance between leveraging semantic importance and preserving representation fidelity.
\begin{table}[H]
\caption{The ablation of $\alpha$ value.}
\label{exp:moralpha}
\centering
\setlength{\tabcolsep}{4pt}
\resizebox{0.88\linewidth}{!}{
    \begin{tabular}{c|ccccc}
    \hline
    \multirow{2}{*}{\textbf{$\alpha$}} &
    \multicolumn{5}{c}{\textbf{Accuracy on Tasks}} \\ \cline{2-6}
    & General$ \uparrow $ & OCR$ \uparrow $ & Video$ \uparrow $ & Audio$ \downarrow $ & NLP$ \uparrow $ \\    \hline
    0.005 & 54.11 & 70.4 & 41.0 & 2.77 & 68.21 \\
    \rowcolor{gray!20}0.01(Ours) & 54.72 & 71.17 & 46.26 & 2.82 & 70.88 \\
    0.05 & 54.53 & 70.93 & 46.11 & 2.79 & 69.78\\
    0.1 & 54.49 & 70.00 & 45.24 & 2.78 & 70.18 \\
    \hline
    \end{tabular}
}
\end{table}

\subsection{Further Analysis}

In this section, we conduct additional analyzes to better understand the behavior and design choices of our model. Specifically, we investigate: (i) the sensitivity to the number of activated experts \(K\) under static v.s. dynamic routing; (ii) the impact of the proportion of virtual experts; and (iii) the effect of scaling the capacity of the token importance estimation module.

\noindent\textbf{Sensitivity to the Number of Activated Experts} We evaluate the robustness of routing strategies under varying expert budgets by analyzing model performance as a function of the average number of activated experts per token (Figure~\ref{fig:expert_sensitivity}). For the static top-$K$, each $K \in \{4,6,8,10\}$ requires a separately trained model. As $K$ decreases, performance degrades substantially across all tasks: general image–text QA accuracy falls by 3.16\% (from 52.55\% at $K$=10 to 49.39\% at $K$=4), OCR accuracy decreases by 2.61\%, and video QA declines by 0.76\%. This underscores the brittleness of static-capacity routing under resource constraints. 

More importantly, AnyExperts demonstrates modality-aware efficiency: under aggressive expert budget reduction, performance degrades gracefully in accordance with the intrinsic information density of each modality. Specifically, for general image–text QA and video QA, reducing \textit{Avg-K} from 8.0 to 4.8 (a 40\% cut) causes negligible performance loss, confirming that visual inputs contain substantial redundancy that can be safely compressed without harming reasoning quality. In contrast, text-dense tasks (OCR and NLP) are more sensitive to capacity reduction. When \textit{Avg-K} is mildly reduced from 8.0 to 7.2 (10\% cut), performance declines only slightly. However, further compression (\textit{e.g.}, Avg-K $< 7.0$) leads to noticeable degradation, as expected for tasks where most tokens carry critical semantic content. 

These results further validate that AnyExperts effectively decouples model capacity from rigid expert allocation, enabling flexible, efficient inference without retraining, especially beneficial for multimodal scenarios where information density varies significantly across modalities.

\begin{table}[H]
\caption{Analysis of virtual expert ratio $\rho_{\max}$ (default: 20\%).}
\label{exp:vexpert}
\centering
\small
\setlength{\tabcolsep}{4pt}
\resizebox{0.95\linewidth}{!}{
    \begin{tabular}{c|ccccc}
    \hline
    \multirow{2}{*}{\textbf{Ratio}} &
    \multicolumn{5}{c}{\textbf{Accuracy on Tasks}} \\ \cline{2-6}
     & General$ \uparrow $ & OCR$ \uparrow $ & Video$ \uparrow $ & Audio$ \downarrow $ & NLP$ \uparrow $ \\ 
    \hline
     10\% & 53.88 & 69.34 & 44.71 & 2.82 & 68.62 \\
     \rowcolor{gray!20} \makecell{20\%(Ours)} & 54.72 & 71.17 & 46.26 & 2.82 & 70.88 \\
     25\% & 53.15 & 70.00 & 45.39 & 2.80 & 57.21 \\
    \hline
    \end{tabular}
}
\end{table}

\noindent\textbf{Impact of Virtual Expert Proportion} We study how the max fraction of virtual experts, controlled by the hyperparameter $\rho_{\max}$, affects model performance. Specifically, we evaluate configurations with $\rho_{\max} = 10\%$, $20\%$, and $25\%$. As shown in Table~\ref{exp:vexpert}, increasing $\rho_{\max}$ from 10\% to 20\% does not degrade performance on any benchmark, and even yields modest gains on several. This suggests that a moderate proportion of virtual experts can enhance model robustness without compromising capacity, as many input tokens exhibit inherent redundancy and do not require full expert computation. However, further increasing $\rho_{\max}$ to 25\% leads to a significant performance drop on NLP tasks, indicating that an excessive share of experts may impair the model’s ability to capture complex linguistic patterns.

\noindent\textbf{Capacity of the Token Importance Estimator} We investigate how model performance varies with the capacity of the token importance estimation network by scaling both its width (hidden dimension) and depth (number of layers). As shown in Table~\ref{exp:tie}, our default lightweight MLP already achieves strong performance across all tasks. Surprisingly, simply widening the MLP ("WideMLP") improves General QA accuracy but degrades performance on Video and OCR tasks, while offering only marginal gains elsewhere, suggesting that excessive capacity may overfit to certain modalities or introduce noise into the importance scores. On the other hand, deepening the network ("DeepMLP") consistently underperforms our default design, likely due to optimization challenges and signal dilution in deeper structures. These results indicate that a compact, well-calibrated importance estimator is not only sufficient but preferable: it provides reliable importance signals without destabilizing routing or incurring unnecessary complexity.
\begin{table}[H]
\caption{Analysis of architecture variants for the token importance estimator.}
\label{exp:tie}
\centering
\setlength{\tabcolsep}{4pt}
\resizebox{0.9\linewidth}{!}{
    \begin{tabular}{c|ccccc}
    \hline
    \multirow{2}{*}{\textbf{Module}} &
    \multicolumn{5}{c}{\textbf{Accuracy on Tasks}} \\ \cline{2-6}
     & General$ \uparrow $ & OCR$ \uparrow $ & Video$ \uparrow $ & Audio$ \downarrow $ & NLP$ \uparrow $ \\
    \hline
    \rowcolor{gray!20} MLP(Ours) & 54.72 & 71.17 & 46.26 & 2.82 & 70.88 \\
     WideMLP & 55.72 & 70.40 & 44.68 & 2.41 & 70.74 \\
     DeepMLP & 53.51 & 70.54 & 44.48 & 2.95 & 68.87 \\
    \hline
    \end{tabular}
}
\end{table}

\subsection{Visualization}

\begin{figure}[!h]
    \centering
    \includegraphics[width=\linewidth]{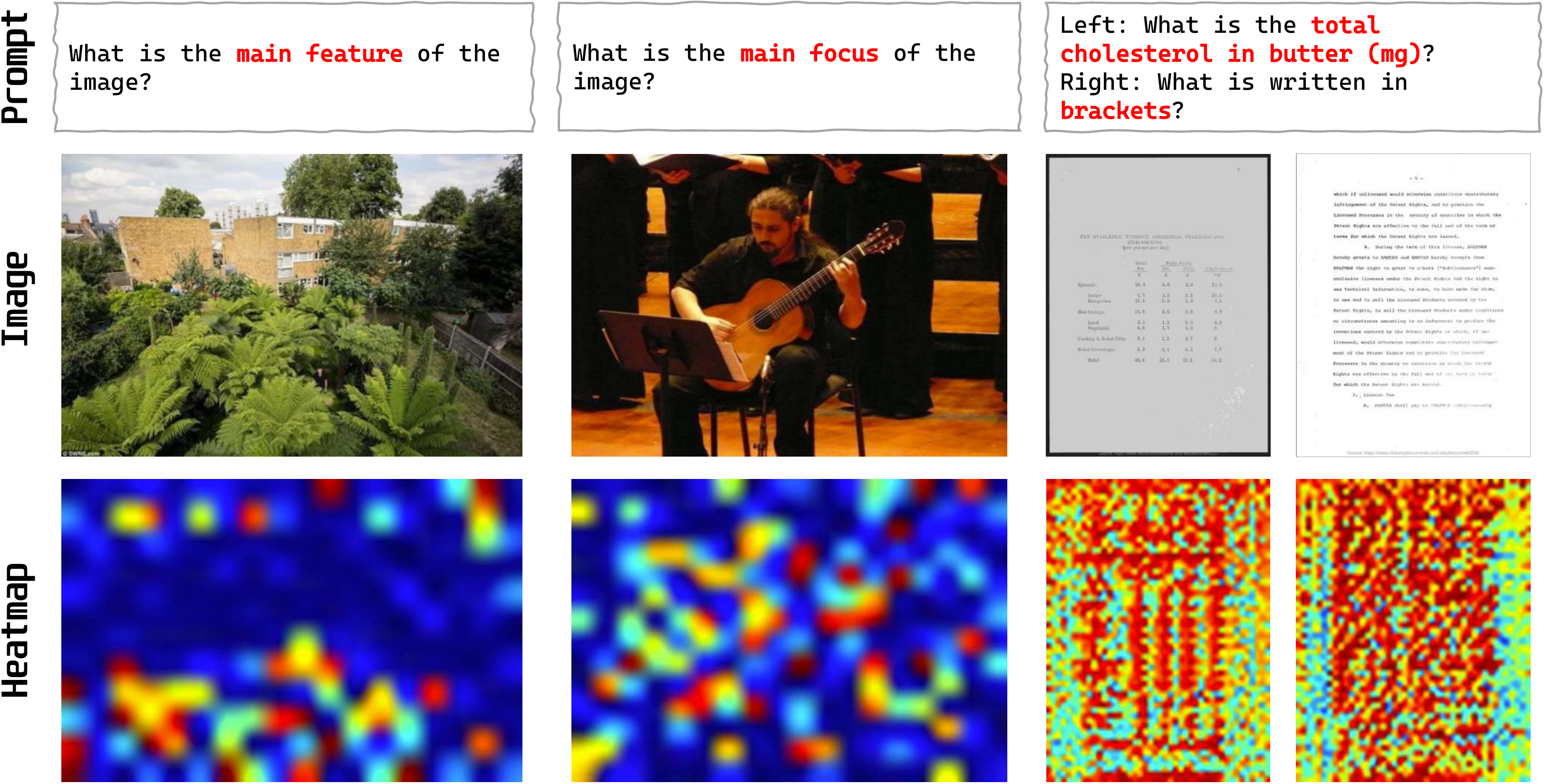}
    \caption{Token importance heatmaps for text–image samples. Redder regions indicate higher importance; bluer regions indicate lower importance.}
    \label{fig:anyexperts_visual_v1}
\end{figure}

\begin{figure}[!h]
    \centering
    \includegraphics[width=1.0\linewidth]{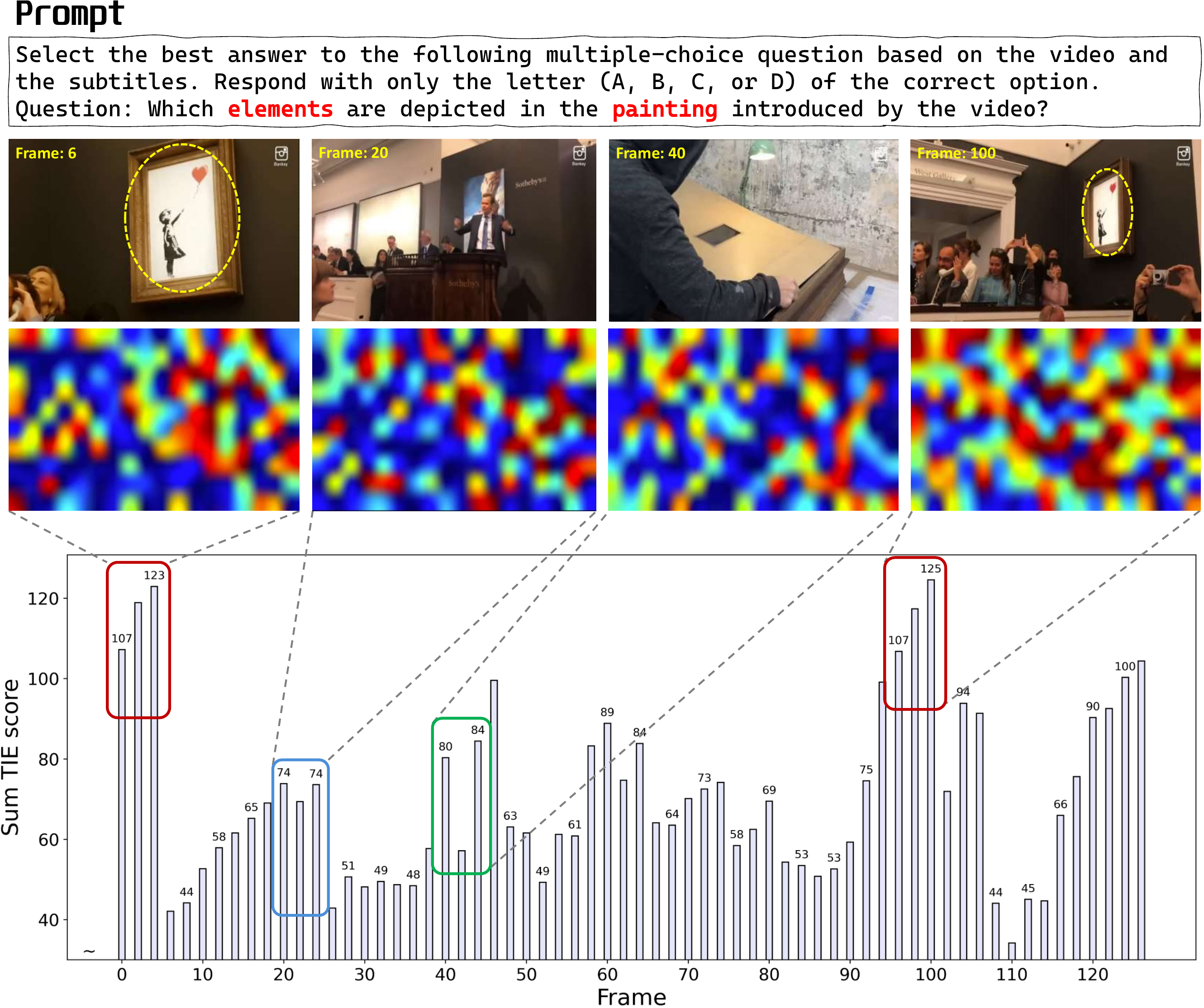}
    \caption{Token importance analysis for text–video samples. Yellow circles indicate answer targets; bar charts show per-frame aggregated token importance over time.}
    \label{fig:anyexperts_visual_v2}
\end{figure}

To illustrate the token importance analysis across modalities, we present heatmap visualizations for text-image and text-video scenarios via Figure~\ref{fig:anyexperts_visual_v1} and Figure~\ref{fig:anyexperts_visual_v2}, respectively. Focusing on image-text samples, Figure~\ref{fig:anyexperts_visual_v1} demonstrates how the model focuses on different foreground objects and environments across various scenarios. Evidently, in the first two samples, the model attends to the extensive garden (as the environment) and the man playing the guitar (as a foreground object). In the latter two samples, it shows extremely high attention to the text-containing regions, while the blank areas receive little focus.

Shifting to video-text, Figure~\ref{fig:anyexperts_visual_v2} employs bar charts to intuitively illustrate the correlation between the sum of token importance weights per frame (vertical) and the frame stream (horizontal). Specifically, the model exhibits heightened attention in the initial few frames, as the video frames contain the target object referenced by the prompt “painting”. Over the subsequent dozens of frames, it remains at a low attention level due to the prolonged absence of the target object. Around the 100th frame, the object reappears and recaptures the model's attention, which is reflected in the corresponding rise in the score sum on the bar chart. These sample visualizations collectively demonstrate the effectiveness of token importance estimation in our model.

\section{Conclusions}
\label{sec:conclusions}
We present AnyExperts, a novel on-demand, budget-aware dynamic routing framework for multimodal Mixture-of-Experts models. By allocating a variable number of expert slots per token based on semantic importance and adaptively balancing real and virtual experts within a strict compute budget, AnyExperts overcomes the inefficiencies of rigid, static-top-$K$ routing. Our approach ensures that semantically rich tokens receive more computational resources, while redundant content is handled efficiently via virtual experts. Extensive experiments across vision, audio, and NLP tasks demonstrate that AnyExperts consistently improves performance under the same compute budget and matches baseline accuracy with fewer expert activations. These results highlight that fine-grained, importance-aware expert allocation is key to unlocking both higher efficiency and stronger performance in multimodal MoE systems.

{
    \small
    \bibliographystyle{ieeenat_fullname}
    \bibliography{main}
}

\clearpage
\setcounter{page}{1}
\maketitlesupplementary

In this appendix, we provide supplementary results and analyses omitted from the main paper due to page constraints. Specifically, Section~\ref{appendix:computation} validates that a 10\% reduction in average activated experts (from 8.0 to 7.2) preserves accuracy across all modalities, as noted in Table 1; Section~\ref{appendix:audioperformance_to_activated_experts} includes the sensitivity analysis for audio understanding, omitted from Figure 3 due to its robustness to expert budget changes; Section~\ref{appendix:per_bench_results} provides per-benchmark performance details underlying the averaged scores reported experiment part;  Finally, Section~\ref{appendix:more_visualization} provides more visualization examples.

\section{Comp-Acc Trade-off}
\label{appendix:computation}
In this section, we present the full performance of AnyExperts on multimodal benchmarks under two expert activation budgets: the default Avg-$K=8.0$ and a reduced Avg-$K=7.2$ (a 10\% decrease). As shown in Table~\ref{tab:our_results}, the model maintains consistently strong results across all modalities (general vision-language, OCR, NLP, speech, and video), when using fewer activated experts.  In most cases, performance remains nearly unchanged, and in some benchmarks (\textit{e.g.}, MMVet, OvOBench), it even slightly improves. The average scores across each modality category show negligible degradation (or minor gains), confirming that AnyExperts achieves high efficiency without sacrificing accuracy. This validates our claim in the main paper that a modest reduction in expert usage preserves overall capability, making the model more computationally economical while retaining robustness.

\section{Audio Robustness}
\label{appendix:audioperformance_to_activated_experts}

In this section, we present a detailed ablation on the number of activated experts for audio understanding, which was omitted from Figure~3 due to its robustness to expert budget variations.  As shown in Table~\ref{exp:topk}, the Word Error Rate (WER) remains largely stable across different activation budgets, both for our dynamic AnyExperts (with average \textit{Avg-$K$}) and static top-$K$ baselines.  This insensitivity suggests that the audio modality in AnyExperts is particularly robust to routing granularity, in contrast to other modalities that exhibit stronger dependence on expert allocation.

\begin{table}[h]
\caption{Ablation on the number of activated experts (Audio-only results). 
Gray rows: our dynamic AnyExperts model with average activation count (\textit{Avg-$K$}). 
White rows: static top-$K$ routing baseline with fixed $K$ per token.}
\label{exp:topk}
\centering
\small
\setlength{\tabcolsep}{8pt}
\begin{tabular}{c c}
\hline
\textbf{Experts / Token} & \textbf{Audio (WER $\downarrow$)} \\
\hline
\multicolumn{2}{l}{\textbf{Dynamic (AnyExperts, Ours)}} \\
\rowcolor{gray!20} \textit{Avg-$K$}=4.8 & 2.99 \\
\rowcolor{gray!20} \textit{Avg-$K$}=5.6 & 2.92 \\
\rowcolor{gray!20} \textit{Avg-$K$}=6.4 & 2.88 \\
\rowcolor{gray!20} \textit{Avg-$K$}=7.2 & 2.82 \\
\rowcolor{gray!20} \textit{Avg-$K$}=8.0 & 2.82 \\
\rowcolor{gray!20} \textit{Avg-$K$}=8.8 & 2.78 \\
\rowcolor{gray!20} \textit{Avg-$K$}=9.6 & 2.78 \\
\hline
\multicolumn{2}{l}{\textbf{Static (Top-$K$ Routing Baseline)}} \\
$K$=4   & 2.98 \\
$K$=6   & 2.82 \\
$K$=8   & 2.84 \\
$K$=10  & 2.78 \\
\hline
\end{tabular}
\end{table}

\begin{table*}[t]
\centering
\small
\setlength{\tabcolsep}{3.3pt}
\caption{Performance of AnyExperts on multimodal benchmarks with default expert activation budget (Avg-$K$=8.0).}
\label{tab:our_results}
\begin{tabular}{l cc | l cc  | l cc | l cc | l cc}
\toprule
\multicolumn{3}{c|}{\textbf{General $\uparrow$}} &
\multicolumn{3}{c|}{\textbf{OCR $\uparrow$}} &
\multicolumn{3}{c|}{\textbf{NLP $\uparrow$}} &
\multicolumn{3}{c|}{\textbf{Speech $\downarrow$}} &
\multicolumn{3}{c}{\textbf{Video $\uparrow$}} \\
\cmidrule(lr){1-3} \cmidrule(lr){4-6} \cmidrule(lr){7-9} \cmidrule(lr){10-12} \cmidrule(lr){13-15}
\textbf{Bench} & \textbf{8.0} & \textbf{7.2} &
\textbf{Bench} & \textbf{8.0} & \textbf{7.2} &
\textbf{Bench} & \textbf{8.0} & \textbf{7.2} &
\textbf{Bench} & \textbf{8.0} & \textbf{7.2} &
\textbf{Bench} & \textbf{8.0} & \textbf{7.2} \\
\midrule

MMBench           & 79.73 & 79.73 & ChartQA            & 83.04 & 82.64 & ARC-C   & 90.17 & 91.19 & Aishell1              & 1.70 & 1.68 & MVBench     & 66.40 & 66.00 \\
AI2D             & 81.19 & 81.19 & TextVQA             & 73.42 & 73.08 & GSM8k   & 84.76 & 84.61 & Aishell2-ios     & 2.78 & 2.78 & VideoMME    & 60.59 & 60.30 \\
MMMU             & 49.11 & 48.67 & OCRBench           & 85.10 & 85.00 & GPQA    & 41.92 & 38.87 & Aishell2-android & 2.76 & 2.78 & OvOBench    & 44.83 & 46.22 \\
MMVet            & 66.15 & 66.97 &                    &       &   &         &       &   & Librispeech-other& 3.01 & 3.00 &             &       &   \\
MathVista        & 66.70 & 66.63 &                    &       &   &         &       &   & Librispeech-clean& 1.38 & 1.38 &             &       &   \\
MMStar           & 60.48 & 60.90 &                    &       &   &         &       &   &                       &      &   &             &       &   \\
\midrule
\rowcolor{gray!20}
Average          & 67.22 & 67.35 & Average            & 80.52 & 80.24 & Average & 72.28 & 71.56 & Average               & 2.32 & 2.32 & Average     & 57.27 & 57.51 \\
\bottomrule
\end{tabular}
\end{table*}

\section{Per-Benchmark Results}
\label{appendix:per_bench_results}

In this section, we provide per-benchmark performance details underlying the averaged scores reported in the main ablation study. The Table~\ref{tab:comparison} includes variants that remove HSM or IAR, as well as sensitivity analyses over hyperparameters $\alpha$, $\beta_{\max}$, and alternative MLP architectures (WildMLP and DeepMLP).

\begin{table*}[t]
\centering
\small
\setlength{\tabcolsep}{2.2pt}
\definecolor{oursbg}{RGB}{230,240,255}      
\definecolor{wohsm_bg}{RGB}{250,220,220}    
\definecolor{woiar_bg}{RGB}{220,235,220}    
\definecolor{alpha_bg}{RGB}{245,240,220}    
\definecolor{beta_bg}{RGB}{250,240,240}     
\definecolor{mlp_bg}{RGB}{235,235,245}      

\caption{Ablation study of AnyExperts. Column groups correspond to: 
\colorbox{oursbg}{\rule{0pt}{6pt}\hspace{7pt}}~Ours (baseline model), 
\colorbox{wohsm_bg}{\rule{0pt}{6pt}\hspace{7pt}}~w/o HSM, 
\colorbox{woiar_bg}{\rule{0pt}{6pt}\hspace{7pt}}~w/o IAR, 
\colorbox{alpha_bg}{\rule{0pt}{6pt}\hspace{7pt}}~$\alpha$-ablation ($\alpha=0.005, 0.05, 0.1$),
\colorbox{beta_bg}{\rule{0pt}{6pt}\hspace{7pt}}~$\beta_{max}$-ablation, and
\colorbox{mlp_bg}{\rule{0pt}{6pt}\hspace{7pt}}~MLP variants, WMLP denote WildMLP and DMLP represent DeepMLP.
$\uparrow$: higher is better; $\downarrow$: lower is better.}
\label{tab:comparison}
\begin{tabular}{
    l
    l
    >{\columncolor{oursbg}}c
    >{\columncolor{wohsm_bg}}c
    >{\columncolor{woiar_bg}}c
    >{\columncolor{alpha_bg}}c
    >{\columncolor{alpha_bg}}c
    >{\columncolor{alpha_bg}}c
    >{\columncolor{beta_bg}}c    
    >{\columncolor{beta_bg}}c    
    >{\columncolor{mlp_bg}}c     
    >{\columncolor{mlp_bg}}c     
}
\toprule
\rowcolor{white}
\textbf{Modality} & \textbf{Benchmark} & \textbf{Ours} & \textbf{w/o HSM} & \textbf{w/o IAR} & \textbf{$\alpha=0.005$} & \textbf{$\alpha=0.05$} & \textbf{$\alpha=0.1$} & $\beta_{\max}=0.1$ & $\beta_{\max}=0.25$ & \textbf{WMLP} & \textbf{DMLP} \\
\midrule

General ($\uparrow$) & MMBench          & 66.84 & 61.17 & 60.13 & 61.08 & 66.15 & 65.12 & 60.91 & 62.63 & 65.89 & 61.00 \\
                     & AI2D             & 67.88 & 65.71 & 66.13 & 65.64 & 68.17 & 68.85 & 67.58 & 66.39 & 68.20 & 67.23 \\
                     & MMMU             & 41.89 & 42.11 & 40.78 & 41.44 & 45.22 & 43.22 & 44.56 & 35.56 & 44.11 & 44.11 \\
                     & MMVet            & 56.38 & 55.09 & 54.77 & 58.12 & 56.24 & 53.90 & 56.15 & 55.87 & 59.45 & 55.37 \\
                     & MathVista        & 49.07 & 49.83 & 47.10 & 51.20 & 47.87 & 48.40 & 48.90 & 52.10 & 50.43 & 49.27 \\
                     & MMStar           & 46.25 & 45.04 & 45.85 & 47.20 & 43.54 & 47.47 & 45.20 & 46.39 & 46.28 & 44.11 \\
\rowcolor{gray!10}
                     & Avg.             & 54.72 & 53.16 & 52.46 & 54.11 & 54.53 & 54.49 & 53.88 & 53.16 & 55.73 & 53.52 \\
\midrule

OCR ($\uparrow$)     & ChartQA          & 73.32 & 71.56 & 73.16 & 72.32 & 72.12 & 73.60 & 71.96 & 72.76 & 72.52 & 74.24 \\
                     & TextVQA          & 66.19 & 64.48 & 64.39 & 65.83 & 66.37 & 64.11 & 63.05 & 65.55 & 65.51 & 64.58 \\
                     & OCRBench         & 74.00 & 72.10 & 72.30 & 73.10 & 74.30 & 72.30 & 73.00 & 71.70 & 73.20 & 72.80 \\
\rowcolor{gray!10}
                     & Avg.             & 71.17 & 69.38 & 69.95 & 70.42 & 70.93 & 70.00 & 69.34 & 70.00 & 70.41 & 70.54 \\
\midrule

Video ($\uparrow$)   & MVBench          & 46.03 & 45.40 & 42.85 & 40.88 & 46.43 & 46.05 & 44.78 & 47.23 & 46.00 & 44.60 \\
                     & VideoMME         & 52.67 & 50.37 & 47.41 & 46.52 & 51.30 & 52.96 & 49.96 & 51.56 & 51.33 & 51.63 \\
                     & OvOBench         & 40.09 & 38.12 & 36.82 & 35.57 & 40.61 & 36.70 & 39.38 & 37.39 & 36.71 & 37.21 \\
\rowcolor{gray!10}
                     & Avg.             & 46.26 & 44.63 & 42.36 & 40.99 & 46.11 & 45.24 & 44.71 & 45.39 & 44.68 & 44.48 \\
\midrule

Speech ($\downarrow$)& Aishell1         & 2.24 & 2.16 & 2.06 & 2.17 & 2.17 & 2.11 & 2.11 & 2.12 & 2.60 & 2.24 \\
                     & Aishell2-ios     & 3.06 & 3.12 & 3.17 & 3.05 & 3.13 & 3.11 & 3.11 & 3.09 & 3.16 & 3.22 \\
                     & Aishell2-android & 3.13 & 3.17 & 3.22 & 3.09 & 3.10 & 3.12 & 3.23 & 3.17 & 3.23 & 3.30 \\
                     & Librispeech-other& 3.83 & 3.81 & 3.87 & 3.72 & 3.78 & 3.77 & 3.88 & 3.79 & 3.80 & 4.04 \\
                     & Librispeech-clean& 1.82 & 1.77 & 1.83 & 1.81 & 1.78 & 1.78 & 1.76 & 1.84 & 1.87 & 1.96 \\
\rowcolor{gray!10}
                     & Avg.             & 2.82 & 2.81 & 2.83 & 2.77 & 2.79 & 2.78 &  2.82 & 2.80 & 2.93 & 2.95 \\
\midrule

NLP ($\uparrow$)     & ARC-C            & 89.15 & 87.80 & 90.51 & 86.10 & 85.76 & 87.12 & 89.83 & 48.14 & 88.81 & 87.46 \\
                     & GSM8k            & 81.58 & 82.18 & 85.29 & 80.67 & 80.36 & 83.02 & 79.68 & 83.09 & 81.50 & 81.80 \\
                     & GPQA             & 41.92 & 36.36 & 33.84 & 37.88 & 42.93 & 40.40 & 36.36 & 40.40 & 41.92 & 37.37 \\
\rowcolor{gray!10}
                     & Avg.             & 70.88 & 68.78 & 69.88 & 68.22 & 69.68 & 70.18 &  68.62 & 57.21 & 70.74 & 68.88 \\
\bottomrule
\end{tabular}
\end{table*}

\section{More Visualization}
\label{appendix:more_visualization}

\begin{figure*}[t]
    \centering
    \includegraphics[width=1.0\linewidth]{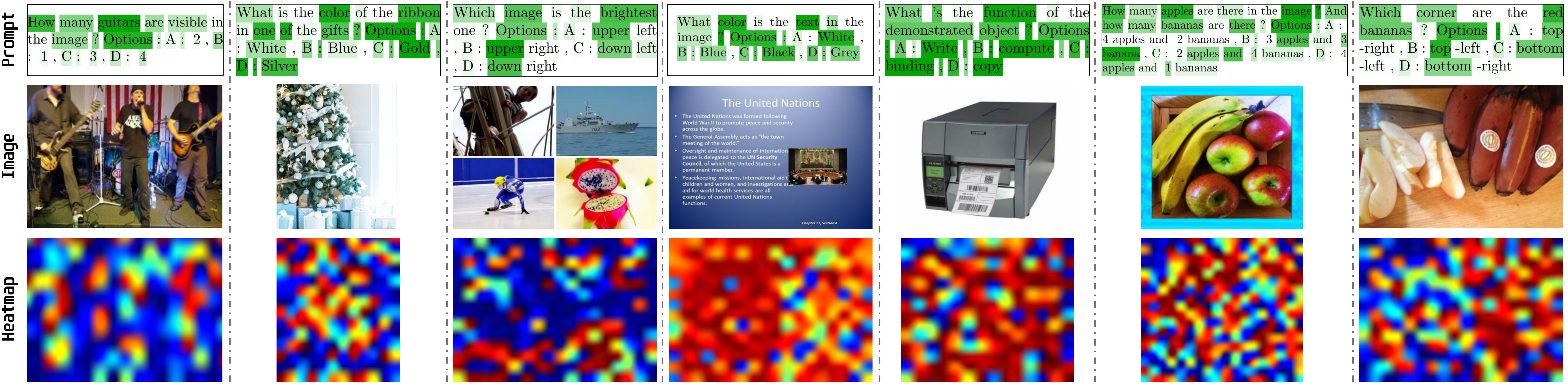}
    \caption{Token importance heatmaps for more text–image samples in general tasks. In heatmaps, redder regions indicate higher importance; bluer regions indicate lower importance. In prompts, areas with deeper green indicate high importance score.}
    \label{fig:supp_vis_01}
\end{figure*}

\begin{figure*}[h]
    \centering
    \includegraphics[width=1.0\linewidth]{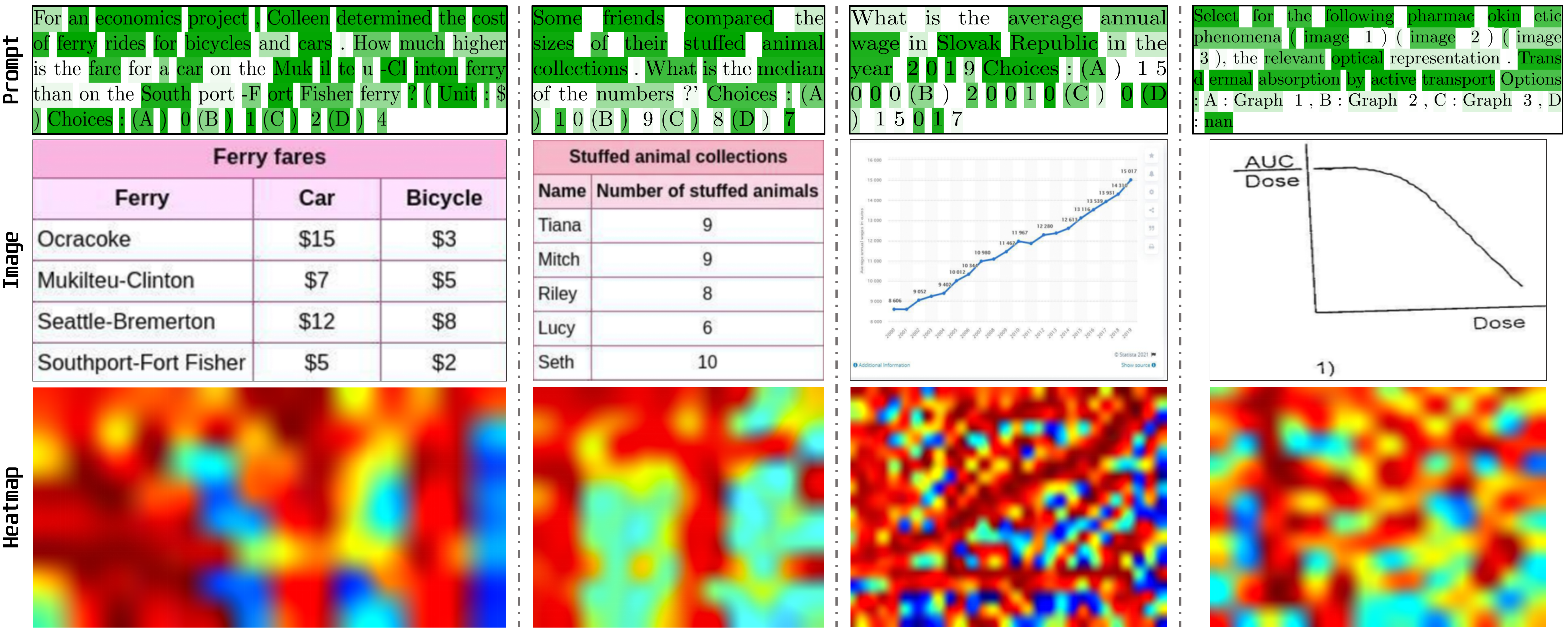}
    \caption{Token importance heatmaps for text–image samples in chart recognition task.}
    \label{fig:supp_vis_02}
\end{figure*}
\begin{figure*}[t]
    \centering
    \includegraphics[width=1.0\linewidth]{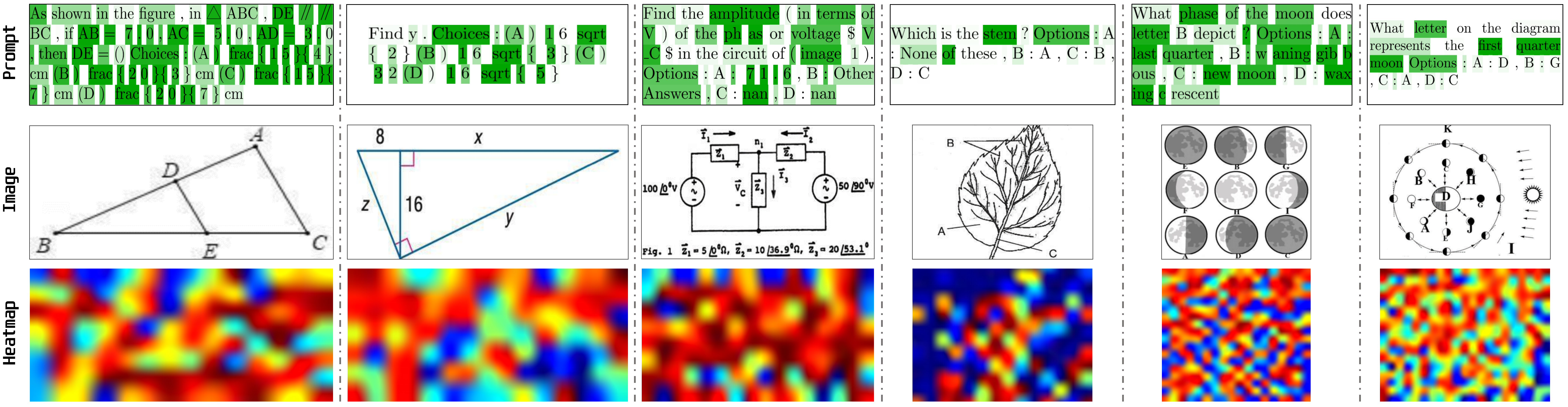}
    \caption{Token importance heatmaps for text–image samples in math and science tasks.}
    \label{fig:supp_vis_03}
\end{figure*}

We performed further visualization analyses on multi-modal data across diverse task types. For the general task in Figure~\ref{fig:supp_vis_01}, we evaluated the model’s performance in target spatial localization, color, position, category recognition, and common sense reasoning. Specifically, in the sample of the first column, the model successfully localized two guitars; in the third column sample, it identified that the brightest regions lie in the two lower sub-figures; and in the sixth column sample, it accurately recognized two bananas and four apples.

\begin{figure}[t]
    \centering
    \includegraphics[width=1.0\linewidth]{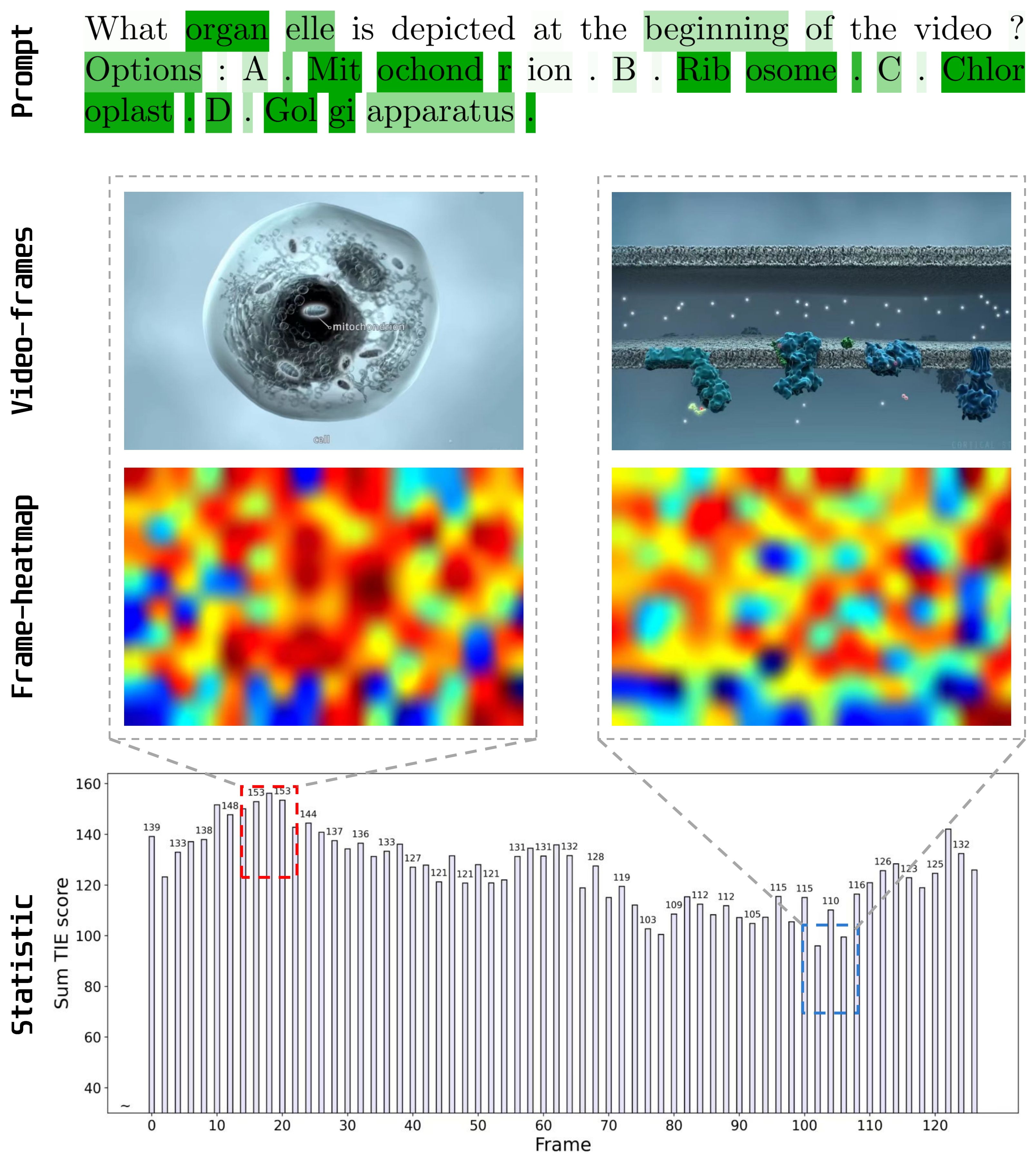}
    \caption{Token importance analysis for text–video samples.}
    \label{fig:supp_vis_04}
\end{figure}

\begin{figure}[h]
    \centering
    \includegraphics[width=1.0\linewidth]{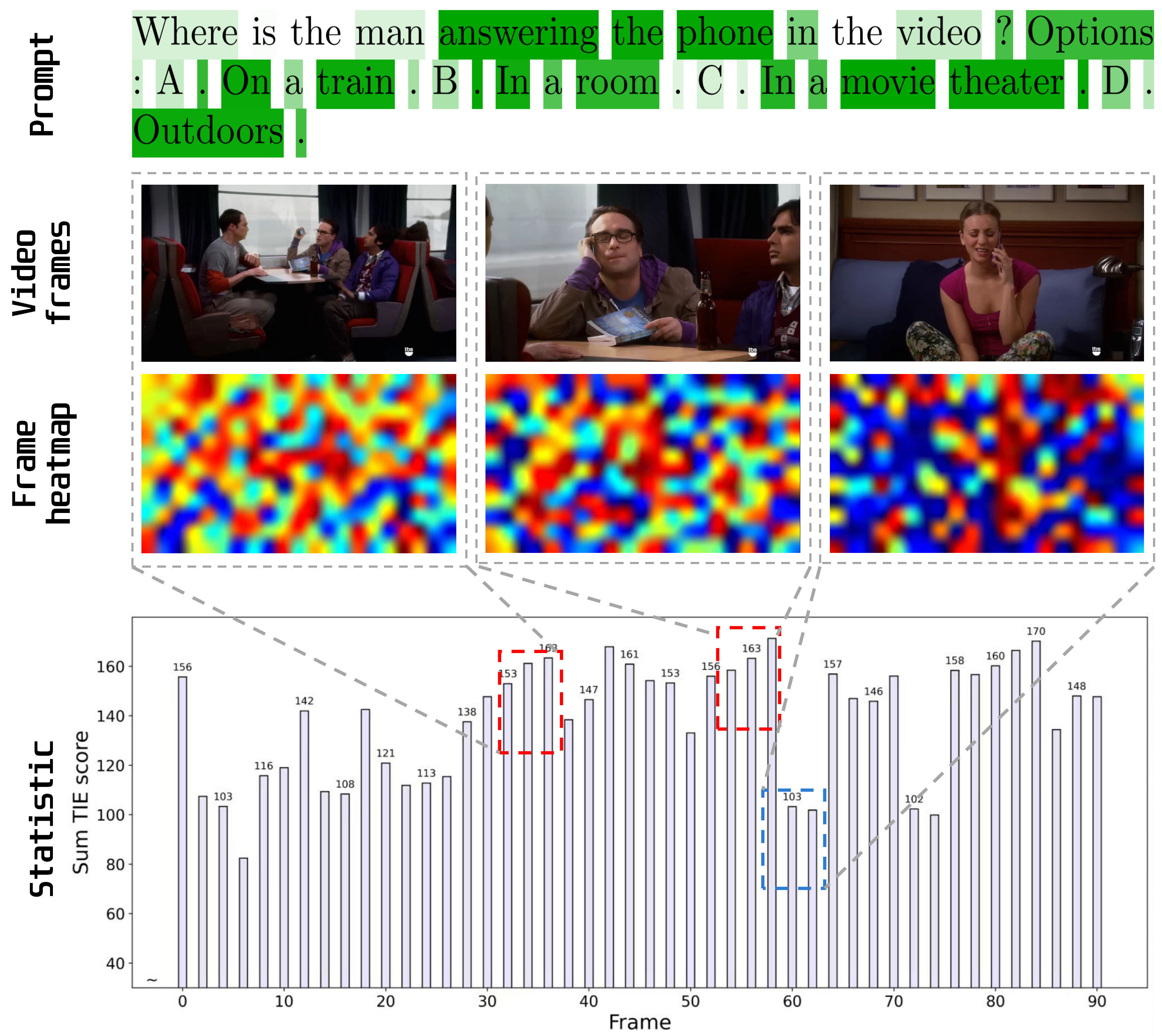}
    \caption{Token importance analysis for text–video samples.}
    \label{fig:supp_vis_05}
\end{figure}

In the chart understanding task (Figure~\ref{fig:supp_vis_02}), the model predominantly concentrated on text-bearing regions of the table in the first two columns of samples, with negligible attention paid to blank areas. For the latter two columns, it effectively captured the extension trend of curves and key text on the coordinate axes.

Regarding the mathematics and science tasks in Figure~\ref{fig:supp_vis_03}, the model attended to the shapes of figures in the first two columns of math samples, prioritizing the positions of vertices and numerical values. In science tasks, it exhibited high attention to the entire circuit image (third column) owing to the sample’s comparative complexity; for the subsequent simpler leaf image, it maintained a sharp focus on the target object. In the final two geography samples, the model noted the number of targets in sub-figures and the process transformations illustrated.

Furthermore, across the three figures above, the model’s attention to different tokens in the text modality is highly discriminative. It shows extremely high attention to terms denoting key targets, while non-critical terms such as "the" and "a" receive generally low attention.

Finally, we present additional video–text examples in Figure~\ref{fig:supp_vis_04} and Figure~\ref{fig:supp_vis_05} that further illustrate our model’s temporal adaptivity. On frames highly relevant to the question, the model assigns higher importance scores and consequently activates significantly more experts; in contrast, less informative frames receive lower importance scores and exhibit reduced average expert activation. This dynamic, content-aware routing enables the model to allocate computational resources more effectively by prioritizing semantically critical information across time.

In summary, it is clear that our model can devote more computational resources to the critical content across different modalities, thus yielding promising performance.

\end{document}